\documentclass[12pt]{article}
\usepackage[margin=1in]{geometry}
\usepackage{microtype}
\usepackage{times}
\usepackage{url}
\usepackage[hidelinks]{hyperref}
\usepackage[utf8]{inputenc}
\usepackage[small]{caption}
\usepackage{graphicx}
\usepackage{amsmath}
\usepackage{amssymb}
\usepackage{amsthm}
\usepackage{booktabs}
\usepackage{float}
\usepackage{algorithm}
\usepackage{algorithmic}
\usepackage[switch]{lineno}
\usepackage{natbib}

\author{
Fran\c{c}ois Pachet$^1$ \quad Jean-Daniel Zucker$^{1,2}$ \\[6pt]
{\normalsize $^1$\,ImagineAllThePeople, Paris, France} \\
{\normalsize $^2$\,IRD, UMMISCO, Sorbonne Universit\'e, France} \\[4pt]
{\normalsize\texttt{pachet@gmail.com, jean-daniel.zucker@ird.fr}}
}

\title{Maximum Entropy Relaxation of Multi-Way Cardinality Constraints for Synthetic Population Generation}

\begin{document}

\maketitle

\begin{abstract}
Generating synthetic populations from aggregate statistics is a core component of microsimulation, agent-based modeling, policy analysis, and privacy-preserving data release. Beyond classical census marginals, many applications require matching heterogeneous unary, binary, and ternary constraints derived from surveys, expert knowledge, or automatically extracted descriptions. Constructing populations that satisfy such multi-way constraints simultaneously poses a significant computational challenge.
We consider populations where each individual is described by categorical attributes and the target is a collection of global frequency constraints over attribute combinations. Exact formulations scale poorly as the number and arity of constraints increase, especially when the constraints are numerous and overlapping.
Grounded in methods from statistical physics, we propose a maximum-entropy relaxation of this problem. Multi-way cardinality constraints are matched in expectation rather than exactly, yielding an exponential-family distribution over complete population assignments and a convex optimization problem over Lagrange multipliers.
We evaluate the approach on NPORS-derived scaling benchmarks with 4 to 40 attributes and compare it primarily against generalized raking. The results show that MaxEnt becomes increasingly advantageous as the number of attributes and ternary interactions grows, while raking remains competitive on smaller, lower-arity instances.

\end{abstract}


\section{Introduction}
Synthetic population generation has become a foundational problem for contemporary AI systems in settings where individual-level microdata are unavailable, legally restricted, or prohibitively expensive to collect. At the same time, many downstream tasks—such as simulation, forecasting, and counterfactual analysis—require large numbers of representative agents rather than aggregate statistics alone.

This need arises across a growing range of domains. In public policy and politics, synthetic populations are used to evaluate reforms, study voting behavior, and explore demographic scenarios without exposing sensitive personal data. In health and epidemiology, they support large-scale simulations of disease propagation, intervention strategies, and healthcare demand. In retail and marketing, they enable customer behavior modeling, demand forecasting, and what-if analyses in environments where individual data are fragmented or protected. More broadly, synthetic populations provide a bridge between aggregate statistics, expert knowledge, and agent-based models, allowing AI systems to reason about societies, markets, and institutions at scale while respecting privacy and regulatory constraints. Recent work on LLM-based synthetic personas further broadens this perspective beyond classical microsimulation: DeepMind's \emph{Persona Generators} expands compact contextual descriptions into diverse synthetic populations tailored to arbitrary settings, with an explicit emphasis on covering rare but consequential trait combinations \citep{Paglieri2026PersonaGenerators}. This renewed interest also aligns with a broader practitioner view that synthetic data can serve as a flexible design lever for shaping AI system behavior.\footnote{For example, Karpathy describes using LLM-generated synthetic conversations to inject identity and behavior into a small language model \citep{Karpathy2025SyntheticData}.}

In these settings, individuals are described by a vector of categorical attributes (e.g., age bracket, gender, region, occupation), and the population must match a set of published or estimated aggregates.
A convenient formalization is to specify a set of frequency constraints over \emph{patterns} of attribute values.
For example: ``$52\%$ female'' (unary), ``$18\%$ female and age 35--44'' (binary), or ``$4\%$ female, age 35--44, and living in \^{I}le-de-France'' (ternary).
Such requirements are naturally expressed in constraint programming as cardinality constraints. Unary cardinalities are efficiently handled by the Global Cardinality Constraint (GCC) with strong filtering \citep{Regin1996}. However, multi-attribute constraints (arity $\ge 2$) are typically modeled as \emph{Sum} constraints over reified predicates or over contingency-table cell counts.
These overlapping linear constraints often provide weak propagation and become difficult to satisfy at scale, especially when many higher-order constraints are specified or when targets are noisy or mutually inconsistent.

Synthetic population generation has been extensively studied in microsimulation, transportation modeling, and survey methodology, using approaches ranging from combinatorial optimization and calibration to iterative reweighting and probabilistic modeling \citep{Beckman1996,VoasWilliamson2000,DemingStephan1940,Fienberg1980}. These methods have enabled large-scale applications, but typically rely on either unary marginals or fully specified contingency tables, and assume internal consistency of the target statistics. In contrast, many contemporary applications involve selectively specified, heterogeneous, and potentially noisy multi-way constraints, for which existing approaches either do not scale or fail to provide robust solutions. This gap motivates alternative formulations that can accommodate partial and higher-order information without requiring exact feasibility.

This paper proposes a probabilistic alternative: a maximum-entropy relaxation of multi-way cardinality-constrained population synthesis.
Instead of searching directly for a discrete population, we infer a distribution over full attribute assignments whose expectations match the required frequencies.
By the maximum-entropy principle \citep{Jaynes1957}, the solution belongs to an exponential family with one parameter per constraint.

This transforms a difficult combinatorial problem into convex optimization over Lagrange multipliers, after which populations can be sampled efficiently. By construction, the maximum-entropy distribution is as unbiased as possible given the imposed constraints.

\subsubsection{Contributions.}
We provide: (i) a formalization of population synthesis constraints as multi-way cardinality (Sum) CSPs over contingency tables; (ii) a maximum-entropy relaxation that recasts this problem as convex optimization over an exponential-family model; (iii) a practical fitting and sampling procedure in the exact-enumeration regime, together with a convex penalized extension for inconsistent targets; and (iv) an NPORS-derived benchmark family that reveals when MaxEnt overtakes generalized raking as dimensionality and constraint arity increase.

\section{Related Work}

\subsection{Cardinality Constraints in CP}
The Global Cardinality Constraint (GCC) \citep{Regin1996} enforces lower and upper bounds on the number of occurrences of atomic values across a set of variables sharing a domain, and supports strong propagation via flow-based filtering. In population synthesis, unary marginals can be expressed directly as GCCs over per-individual attribute variables. However, GCCs do not naturally extend to multi-attribute patterns: binary and ternary marginals correspond to joint counts rather than counts of atomic values, and therefore cannot be captured by a single GCC over the original variables. Such constraints typically require reification and additional Sum constraints, or contingency-table formulations with explicit count variables, which substantially weakens propagation. 
Recent work by \citet{PetitPachot2025} proposes a CP framework for exact synthetic population generation from aggregate statistics and structural relations, using a batch-based construction strategy rather than a single global solve. 
The paper provides a limited extension-preserving result for the marginal-distribution objective in isolation, but it does not prove global completeness or global optimality for the full model once horizontal constraints, diversity, joint features, time limits, and irrevocable batch commitments are included.
Our work targets the same Sum-based regime, but replaces exact feasibility by a probabilistic relaxation and optimizes a single global model over all constraints at once; it should therefore be viewed as complementary to these encodings rather than as a replacement for classical GCC propagation.

\subsection{Contingency Tables, IPF, and Log-Linear Models}

Fitting contingency tables from marginal distributions has a long tradition in statistics, notably through iterative proportional fitting (IPF) and log-linear models \citep{DemingStephan1940,Bishop1967,Fienberg1980}.
When only first-order marginals are specified, IPF is known to converge to the unique maximum-entropy distribution consistent with these constraints, and can be viewed as a special case of maximum-likelihood estimation in log-linear models with main effects only \citep{Csiszar1975,DarrochRatcliff1972}.

However, this equivalence fundamentally breaks down when constraints involve multi-attribute patterns.
Enforcing binary or ternary marginals using IPF would require operating on full contingency tables whose dimensionality grows exponentially with the number of attributes \citep{Fienberg1980}.
In realistic population synthesis settings, only a sparse and heterogeneous subset of multi-way marginals is available, which prevents the construction of such tables in practice \citep{VoasWilliamson2000,LomaxNorman2016}.

Several extensions of raking have been proposed to incorporate additional constraints, including generalized regression estimators and iterative proportional updating (IPU) \citep{DevilleSarndal1992,Ye2009IPU}.
These approaches rely on heuristic weight adjustments and auxiliary interaction variables, and do not provide general guarantees of feasibility or convergence when constraints are overlapping or selectively specified \citep{ChambersSkinner2003,LomaxNorman2016}.
As a result, IPF and its generalizations cannot serve as principled baselines for population synthesis under arbitrary multi-way cardinality constraints.

\subsection{Bayesian Networks and Probabilistic Graphical Models}
Bayesian networks represent joint distributions through local conditional dependencies structured as directed acyclic graphs \citep{Pearl1988}.
While effective for probabilistic inference under assumed causal or dependency structures, they are not well suited for enforcing exact or near-exact global cardinality constraints over multi-variable configurations.
Encoding such constraints within a Bayesian network would require auxiliary counting variables or high-order dependencies spanning the entire population, thereby eliminating the tractability advantages of Bayesian factorization and rendering inference computationally prohibitive \citep{Cooper1990}.
Moreover, Bayesian networks do not guarantee the preservation of observed marginal distributions, whereas maximum-entropy models explicitly enforce aggregate constraints through convex optimization.
The approach proposed here is therefore better understood as a constraint-driven distribution construction method rather than as a probabilistic inference problem amenable to standard Bayesian network formulations.
\subsection{Positioning our work}

Our approach directly targets the regime where constraints are expressed as sums over arbitrary attribute patterns, including unary, binary, and ternary marginals.
Rather than enumerating full contingency tables or relying on heuristic reweighting schemes, we formulate population synthesis as a maximum-entropy problem under linear expectation constraints.

In this perspective, IPF appears as a degenerate special case of our framework restricted to unary marginals.
Our contribution should therefore be viewed as extending classical population synthesis techniques to the Sum-based constraint regime that arises in realistic high-dimensional settings.
\section{Problem Formulation}
\label{sec:problemformulation}
\subsection{Attribute Space and Populations}
Let $K$ be the number of categorical attributes, with finite domains $D_1,\dots,D_K$.
A persona is a full assignment
\[
x = (a_1,\dots,a_K) \in \mathcal{X} \triangleq D_1 \times \cdots \times D_K,
\]
and $|\mathcal{X}| = \prod_{k=1}^K |D_k|$.
A population of size $N$ can be represented either as a multiset of individuals $\{x^{(1)},\dots,x^{(N)}\}$ or, equivalently, as a contingency table of nonnegative integer counts
\[
C = (C_x)_{x\in\mathcal{X}}, \qquad C_x \in \mathbb{Z}_{\ge 0}, \qquad \sum_{x\in\mathcal{X}} C_x = N.
\]

\subsection{Multi-Way Cardinality Constraints (Sum-CSP)}
We consider constraints specified over \emph{patterns} (subsets) $\mathcal{S}_j \subseteq \mathcal{X}$ with target frequency $\alpha_j\in[0,1]$ (or target count $M_j=\alpha_j N$).
Each pattern induces an indicator feature
\[
f_j(x) \triangleq \mathbf{1}[x\in \mathcal{S}_j].
\]
In a strict CSP over contingency-table variables, the constraints take the form
\begin{equation}
\sum_{x\in \mathcal{S}_j} C_x = M_j, \qquad j=1,\dots,m,
\label{eq:sumcsp}
\end{equation}
together with nonnegativity and the total-size constraint.
When $\mathcal{S}_j$ corresponds to fixing 1, 2, or 3 attributes, \eqref{eq:sumcsp} encodes unary, binary, or ternary marginals respectively.
As arity and overlap increase, \eqref{eq:sumcsp} becomes difficult for CP/MIP solvers, and feasibility may fail under inconsistent targets.

\subsection{Computational Hardness and Motivation for Relaxation}
Problem \eqref{eq:sumcsp} defines a system of linear equalities over $|\mathcal{X}|$ integer variables, where $|\mathcal{X}|$ grows exponentially with the number of attributes $K$.
Even for moderate $K$, explicitly representing all contingency-table cells is infeasible.

Moreover, the constraints in \eqref{eq:sumcsp} are highly overlapping: each variable $C_x$ typically participates in many pattern constraints, and the resulting constraint matrix is dense and ill-conditioned.
As a consequence, exact feasibility is fragile: small inconsistencies in the target marginals may render the system infeasible.

From a computational perspective, solving \eqref{eq:sumcsp} exactly amounts to finding a nonnegative integer solution to a large-scale multi-way marginal consistency problem, which is known to be NP-hard in general and intractable for standard CP or MIP solvers beyond small instances.
Even when feasibility holds, the space of solutions is typically extremely large, making arbitrary solution selection undesirable.

These limitations motivate the use of relaxed formulations that trade exact satisfaction of constraints for probabilistic or entropic consistency, while preserving the essential structure of the marginal information.

\section{Maximum Entropy Relaxation}

Maximum entropy modeling under linear expectation constraints is equivalent to exponential-family modeling: the MaxEnt solution admits a canonical exponential form whose sufficient statistics correspond to the constrained features. While exponential-family and energy-based models are typically learned from microdata via likelihood-based objectives \citep{Wainwright2008}, our setting departs from this paradigm by inferring parameters directly from aggregate multi-way cardinality constraints. The resulting model is therefore used as an approximate solver and sampler for large-scale combinatorial population constraints, rather than as a density estimator over observed samples.

The maximum entropy principle provides a principled way to construct probability distributions from partial information, originally introduced in statistical physics to characterize equilibrium distributions under macroscopic constraints \citep{Jaynes1957}. Its central idea is to select, among all distributions satisfying the known constraints, the one that makes the fewest additional assumptions. Beyond physics, maximum entropy models have been successfully applied in machine learning and structured data modeling to capture higher-order statistical regularities while avoiding overfitting \citep{Sakellariou2017}. In our context, this principle yields a global optimization-based relaxation of multi-way cardinality-constrained population synthesis and suggests a natural penalized extension for noisy or mutually inconsistent targets.

\subsection{Expectation Constraints}
We replace the integer table $C$ by a distribution $p$ over $\mathcal{X}$ and enforce constraints in expectation:
\begin{equation}
\mathbb{E}_p[f_j] = \sum_{x\in\mathcal{X}} p(x)\, f_j(x) = \alpha_j, \qquad j=1,\dots,m.
\label{eq:expconstraints}
\end{equation}
This is a convex relaxation of \eqref{eq:sumcsp}, replacing exact counts by expected frequencies.

\subsection{Soft Constraints for Inconsistent Targets}
When the target vector $\alpha$ lies outside the moment polytope generated by the features $\{f_j\}_{j=1}^m$, the equalities in \eqref{eq:expconstraints} cannot all be satisfied simultaneously. A natural soft alternative is therefore to optimize directly over distributions:
\begin{equation}
\min_{p \in \Delta(\mathcal{X})}
\; -H(p) + \frac{\beta}{2} \sum_{j=1}^m w_j \left(\mathbb{E}_p[f_j] - \alpha_j\right)^2,
\label{eq:softmaxent}
\end{equation}
where $\beta > 0$ controls the trade-off between entropy and constraint fidelity and $w_j \ge 0$ optionally rescales heterogeneous constraints. This objective is convex in $p$ because negative entropy is convex and each penalty term is a square of a linear functional. When the hard constraints are feasible and $\beta$ is large, the solution approaches the standard MaxEnt model. In the present paper, however, the empirical evaluation focuses on the hard-constraint regime induced by empirical marginals extracted from observed populations; the penalized formulation in \eqref{eq:softmaxent} is included to make the treatment of noisy targets explicit.

\subsection{MaxEnt Principle and Exponential Family}
Among distributions satisfying \eqref{eq:expconstraints}, we choose the one with maximum Shannon entropy:
\[
\max_{p \in \Delta(\mathcal{X})} \; H(p)
\quad \text{s.t.}\quad
\mathbb{E}_p[f_j]=\alpha_j \;\; \forall j,
\]
where $H(p)=-\sum_x p(x)\log p(x)$.
By classical duality results \citep{Jaynes1957,Wainwright2008}, if the constraint set is feasible, the unique maximizer has the exponential-family form
\begin{equation}
p_\lambda(x)=\frac{1}{Z(\lambda)}\exp\!\left(\sum_{j=1}^m \lambda_j f_j(x)\right),
\label{eq:expfamily}
\end{equation}
where $\lambda\in\mathbb{R}^m$ are Lagrange multipliers and $Z(\lambda)=\sum_x \exp(\sum_j \lambda_j f_j(x))$ is the partition function.

\subsection{Dual Objective and Convexity}
The dual problem is
\begin{equation}
\min_{\lambda\in\mathbb{R}^m} \;\; \Phi(\lambda)\triangleq \log Z(\lambda) - \sum_{j=1}^m \lambda_j \alpha_j.
\label{eq:dual}
\end{equation}
Its gradient is
\[
\nabla \Phi(\lambda) = \mathbb{E}_{p_\lambda}[f] - \alpha,
\]
and the Hessian equals the covariance matrix $\mathrm{Cov}_{p_\lambda}(f)$, implying convexity and, under mild conditions, strict convexity \citep{Wainwright2008}.

\section{Optimization and Sampling}
\subsection{Learning \texorpdfstring{$\lambda$}{lambda}}
We learn $\lambda$ by minimizing \eqref{eq:dual} using gradient-based optimization.
At each iteration, we generate samples from the current model by running an MCMC scheme with Metropolis updates targeting the Gibbs distribution defined by the current parameters. The empirical moments computed from these sampled populations are then used to update $\lambda$.
The overall procedure is summarized in Algorithm~\ref{alg:maxent}.

\begin{algorithm}[t]
\caption{MaxEnt fitting from multi-way cardinality targets}
\label{alg:maxent}
\begin{algorithmic}[1]
\REQUIRE Attribute space $\mathcal{X}$; patterns $\{\mathcal{S}_j\}_{j=1}^m$; targets $\alpha\in[0,1]^m$
\STATE Initialize $\lambda \leftarrow 0$
\REPEAT
    \STATE Compute $p_\lambda(x)$ via \eqref{eq:expfamily} (exact or approximate normalization)
    \STATE Compute $\hat{\alpha}_j \leftarrow \mathbb{E}_{p_\lambda}[f_j] = \sum_x p_\lambda(x)\mathbf{1}[x\in\mathcal{S}_j]$
    \STATE Update $\lambda \leftarrow \lambda - \eta \, (\hat{\alpha}-\alpha)$ \COMMENT{(e.g., gradient descent; use L-BFGS in practice)}
\UNTIL{convergence}
\RETURN $\lambda$
\end{algorithmic}
\end{algorithm}

\subsection{Generating a Population}
Once $\lambda$ is learned, we sample a population of size $N$ i.i.d.:
\[
x^{(1)},\dots,x^{(N)} \sim p_\lambda.
\]
Empirical frequencies concentrate around targets by standard concentration inequalities.
If exact integer marginals are required, a lightweight \emph{repair} step (e.g., local swaps) can be applied to match selected constraints while minimally perturbing others (left for future work).

\section{Experimental Evaluation}
\label{sec:evaluation}
We evaluate the proposed Maximum Entropy relaxation on a family of NPORS-derived scaling benchmarks with ternary-or-less cardinality constraints. The goal is to isolate how relative performance changes with the number of attributes, the constraint arity, and the sampled population size.

\subsection{Experimental Setup}
\label{sec:setup}

To obtain realistic yet controllable instances, we use the 2024 National Public Opinion Reference Survey (NPORS) as a source population. After preprocessing, the dataset contains $N = 5{,}626$ respondents and several hundred categorical or discretized variables with explicit missing-value codes.

We construct five NPORS-derived problems by selecting subsets of $K \in \{4,12,20,28,40\}$ categorical attributes with moderate domain sizes, clear semantic interpretation, and limited structural missingness. Each subset induces an empirical population in the formalism of Section~\ref{sec:problemformulation}, from which we extract global cardinality constraints over attribute patterns of increasing arity:
\begin{itemize}
\item \textbf{Unary constraints}, corresponding to marginal distributions of single attributes;
\item \textbf{Binary constraints}, corresponding to joint marginals over pairs of attributes;
\item \textbf{Ternary constraints}, corresponding to joint marginals over triples of attributes.
\end{itemize}

Unary marginals are always included in full. For $K < 20$, all candidate pairs and triples are retained. For $K \ge 20$, binary and ternary interactions are ranked by informativeness and truncated to the top 50 pairs and top 50 triples, respectively. This keeps the benchmark family comparable across problem sizes while allowing the number of atomic constraints to reflect domain size and interaction complexity.
  
\subsubsection{Inputs.}
The construction algorithm takes as input:
\begin{itemize}
\item a population dataset $C = (C_x)_{x \in \mathcal{X}}$;
\item a binary constraint budget $n_2 \in \mathbb{N}^+$ or rate $\rho_2 \in (0,1]$,
      controlling the number of variable pairs retained;
\item a ternary constraint budget $n_3 \in \mathbb{N}^+$ or rate $\rho_3 \in (0,1]$,
      controlling the number of variable triples retained.
\end{itemize}
Unary constraints are always extracted in full (one constraint per attribute).

\subsubsection{Variable selection by informativeness.}
Rather than selecting pairs and triples arbitrarily, we rank them by their statistical
informativeness and retain the top-$k$ most informative ones:
\begin{itemize}
\item \textbf{Binary pairs} are ranked by \emph{Normalized Mutual Information} (NMI)
      computed on the joint distribution of each pair in the population;
\item \textbf{Ternary triples} are ranked by \emph{KL divergence} between the observed
      joint marginal and the MaxEnt-IPF reference distribution that matches all three
      pairwise marginals of the triple.
\end{itemize}
Given a budget $n_2$ (resp.\ $n_3$), the top-$n_2$ pairs (resp.\ top-$n_3$ triples)
are retained.
When a rate $\rho_2$ (resp.\ $\rho_3$) is specified instead,
$k = \lceil \rho \times |\text{candidates}| \rceil$ pairs (resp.\ triples) are selected.

\subsubsection{Output.}
The algorithm outputs a constraint problem defined by a set $\mathcal{C}$ of global Sum
constraints enforcing exact pattern counts for the selected marginals.
The total number of atomic constraints equals the sum of the number of observed values
across all selected attributes, pairs, and triples --- which may be orders of magnitude
larger than the number of selected variable pairs or triples
(e.g., 50 triples yielding up to 17\,883 atomic constraints for a 40-variable problem).

\subsubsection{Constraint Extraction Algorithm.}
The construction procedure is summarized in Algorithm~\ref{alg:constraint-extraction}.
Unary, binary, and ternary marginals are extracted from the empirical population and included according to the specified budgets or rates.

\begin{algorithm}[htbp]
\caption{Extraction of Reference Constraints from a Population Dataset}
\label{alg:constraint-extraction}
\begin{algorithmic}[1]
\REQUIRE Population $C$;
         binary budget $n_2 \in \mathbb{N}^+$ \textbf{or} rate $\rho_2 \in (0,1]$;
         ternary budget $n_3 \in \mathbb{N}^+$ \textbf{or} rate $\rho_3 \in (0,1]$
\ENSURE Constraint set $\mathcal{C}$
\STATE $\mathcal{C} \leftarrow \emptyset$

\STATE \textbf{--- Phase 1: Unary constraints} \COMMENT{all attributes, always included in full}
\FOR{each attribute $A_i$, $i = 1,\dots,K$}
    \STATE compute unary marginal of $A_i$ over $C$
    \STATE add one Sum constraint per observed value of $A_i$ to $\mathcal{C}$
\ENDFOR

\STATE \textbf{--- Phase 2: Binary constraints} \COMMENT{top-$k_2$ pairs by Normalized Mutual Information (NMI)}
\FOR{each attribute pair $(A_i, A_j)$ with $i < j$}
    \STATE $s_{ij} \leftarrow \mathrm{NMI}(A_i, A_j)$
\ENDFOR
\STATE $k_2 \leftarrow n_2$ \textbf{if} $n_2$ given, \textbf{else} $\lceil \rho_2 \cdot \tbinom{K}{2} \rceil$
\FOR{each of the $k_2$ top-ranked pairs $(A_i, A_j)$ by $s_{ij}$ \textbf{(descending)}}
    \STATE compute binary marginal of $(A_i, A_j)$ over $C$
    \STATE add one Sum constraint per observed value combination to $\mathcal{C}$
\ENDFOR

\STATE \textbf{--- Phase 3: Ternary constraints} \COMMENT{top-$k_3$ triples by Kullback--Leibler (KL) divergence}
\FOR{each attribute triple $(A_i, A_j, A_k)$ with $i < j < k$}
    \STATE $\hat{P}_{ijk} \leftarrow \mathrm{IPF}(P_{ij},\, P_{ik},\, P_{jk})$
           \COMMENT{Iterative Proportional Fitting: MaxEnt dist.\ matching all pairwise marginals}
    \STATE $s_{ijk} \leftarrow \mathrm{KL}(P_{ijk} \,\|\, \hat{P}_{ijk})$
\ENDFOR
\STATE $k_3 \leftarrow n_3$ \textbf{if} $n_3$ given, \textbf{else} $\lceil \rho_3 \cdot \tbinom{K}{3} \rceil$
\FOR{each of the $k_3$ top-ranked triples $(A_i, A_j, A_k)$ by $s_{ijk}$ \textbf{(descending)}}
    \STATE compute ternary marginal of $(A_i, A_j, A_k)$ over $C$
    \STATE add one Sum constraint per observed value combination to $\mathcal{C}$
\ENDFOR

\RETURN $\mathcal{C}$
\end{algorithmic}
\end{algorithm}
\begin{table}[H]
\centering
\small
\begin{tabular}{lrrrr}
\toprule
\textbf{Problem} & \textbf{Unary} & \textbf{Binary} & \textbf{Ternary} & \textbf{Total} \\
\midrule
4-Var  & 12  & 54     & 92      & 158     \\
12-Var & 35  & 117    & 1\,082   & 1\,234   \\
20-Var & 67  & 510    & 2\,249   & 2\,826   \\
28-Var & 107 & 703    & 3\,824   & 4\,634   \\
40-Var & 197 & 1\,329  & 17\,883  & 19\,409  \\
\bottomrule
\end{tabular}
\caption{Number of nonzero atomic cells induced by the selected unary, binary, and ternary marginals in the NPORS-derived scaling benchmarks ($N=5{,}626$ in the source population). Reported counts correspond to observed support sizes summed over retained marginals, not simply to the number of variables, pairs, or triples. For $K \geq 20$, the numbers of retained binary and ternary marginals are capped at 50.}
\label{tab:nb-constraints}
\end{table}

This procedure yields the five NPORS-derived scaling benchmarks used throughout the paper. Table~\ref{tab:nb-constraints} reports the number of atomic constraints, defined as the observed nonzero cells contributed by the retained unary, binary, and ternary marginals, rather than the number of marginals themselves. The reported values are therefore obtained by summing the observed support sizes of all retained marginals within each arity. In this setting, the support size of a marginal is the number of category combinations that occur at least once in the source population. For example, the ternary marginal \texttt{BORN-DEVICE1A-EMINUSE}\footnote{\texttt{BORN} indicates whether the respondent identifies as a born-again or evangelical Christian, \texttt{DEVICE1A} indicates cell phone ownership, and \texttt{EMINUSE} indicates whether the respondent uses the internet or email.} involves three variables with three observed levels each, so its full Cartesian product contains \(3\times 3\times 3=27\) possible cells; however, only 22 of these combinations are observed in the NPORS source population, and its support size is therefore 22. The 4-Var benchmark makes this accounting explicit: the four variables contribute \(4\times 3=12\) unary constraints, the six retained pairs each have full \(3\times 3\) support and thus contribute \(6\times 9=54\) binary constraints, and the four retained triples have support sizes 22, 23, 25, and 22, which sum to 92 ternary constraints. The total for 4-Var is thus \(12+54+92=158\). The same principle explains the larger counts in higher-dimensional benchmarks: for instance, the 12-Var benchmark contains 12 unary marginals but 35 unary atomic constraints, 14 binary marginals but 117 binary atomic constraints, and 44 ternary marginals but 1,082 ternary atomic constraints. More generally, the ternary count grows rapidly with both the number of retained triples and the attribute domain sizes, reaching 17,883 atomic constraints in the 40-Var benchmark despite the cap of 50 retained triples.


\subsubsection{Compared methods.}
The main empirical comparison is between MaxEnt and generalized raking. We focus on this comparison because both methods produce approximate synthetic populations under the same selected marginals and remain meaningful across all benchmark sizes. Exact CP encodings motivate the relaxation but are not used as the primary quantitative baseline in the present draft.

\subsubsection{OR-Tools CP-SAT formulation.}
For completeness, we also implemented an exact reference model with OR-Tools CP-SAT. Let $V$ denote the set of selected categorical variables and let $n$ be the desired size of the synthetic population. For each variable $v \in V$, the domain $\mathcal{D}_v$ is the set of categories observed in the reference data, with missing-value codes retained as ordinary categories. From the reference population, we compute all unary marginals and, up to a user-defined maximum arity $k \in \{1,2,3\}$, all pairwise and ternary marginals. For each scope $S \subseteq V$ such that $1 \leq |S| \leq k$, and for each observed assignment $a \in \prod_{v \in S} \mathcal{D}_v$, we define a target count $t_{S,a}$. When the requested synthetic size differs from the source population size, these counts are proportionally rescaled so that they sum to $n$. The CP-SAT model then introduces integer decision variables $x_{i,v}$ encoding the category assigned to variable $v$ for synthetic individual $i$, together with auxiliary count variables
\[
y_{S,a} = \sum_{i=1}^{n} \mathbf{1}[x_{i,S} = a]
\]
for every stored marginal cell. Depending on the experimental setting, the model either enforces $y_{S,a} = t_{S,a}$ exactly (hard mode) or minimizes the $\ell_1$ discrepancy
\[
\sum_{S,a} \left| y_{S,a} - t_{S,a} \right|.
\]
Only marginal cells with nonzero empirical support are stored in the generated instance; consequently, unobserved cells are not encoded as structural zeros and remain unconstrained. In the present draft, this CP-SAT model is included as an exact reference formulation, while the systematic quantitative comparison focuses on MaxEnt and generalized raking.

\subsubsection{Implementation details.}
All MaxEnt models are fitted in the exact-enumeration regime: expectations over $\mathcal{X}$ are computed explicitly and the parameters $\lambda$ are optimized with L-BFGS. Generalized raking follows a standard iterative proportional fitting procedure: each individual carries a positive weight, initialized uniformly, and constraints are processed sequentially; for each constraint, the weights of matching individuals are multiplicatively rescaled so that the weighted sum matches the target value. The procedure is iterated 1\,000 times.

\subsubsection{Metric.}
For a sampled population, let $\hat{\alpha}_j$ denote the empirical frequency associated with constraint $j$. We report the mean relative constraint error
\begin{equation}
\mathrm{MRE} \triangleq \frac{1}{m}\sum_{j=1}^m \frac{|\hat{\alpha}_j - \alpha_j|}{\alpha_j},
\label{eq:mre}
\end{equation}
together with runtime. Because only marginal cells with nonzero empirical support are retained, all target frequencies $\alpha_j$ are strictly positive.

\subsection{Results on NPORS-Derived Scaling Benchmarks}
\label{sec:npors}

All results below use the five NPORS-derived problems described above. We vary the sampled population size $N$ to study how each method behaves under increasing sampling pressure while keeping the underlying constraint system fixed.

\begin{figure}[H]
  \centering
  \includegraphics[width=\textwidth]{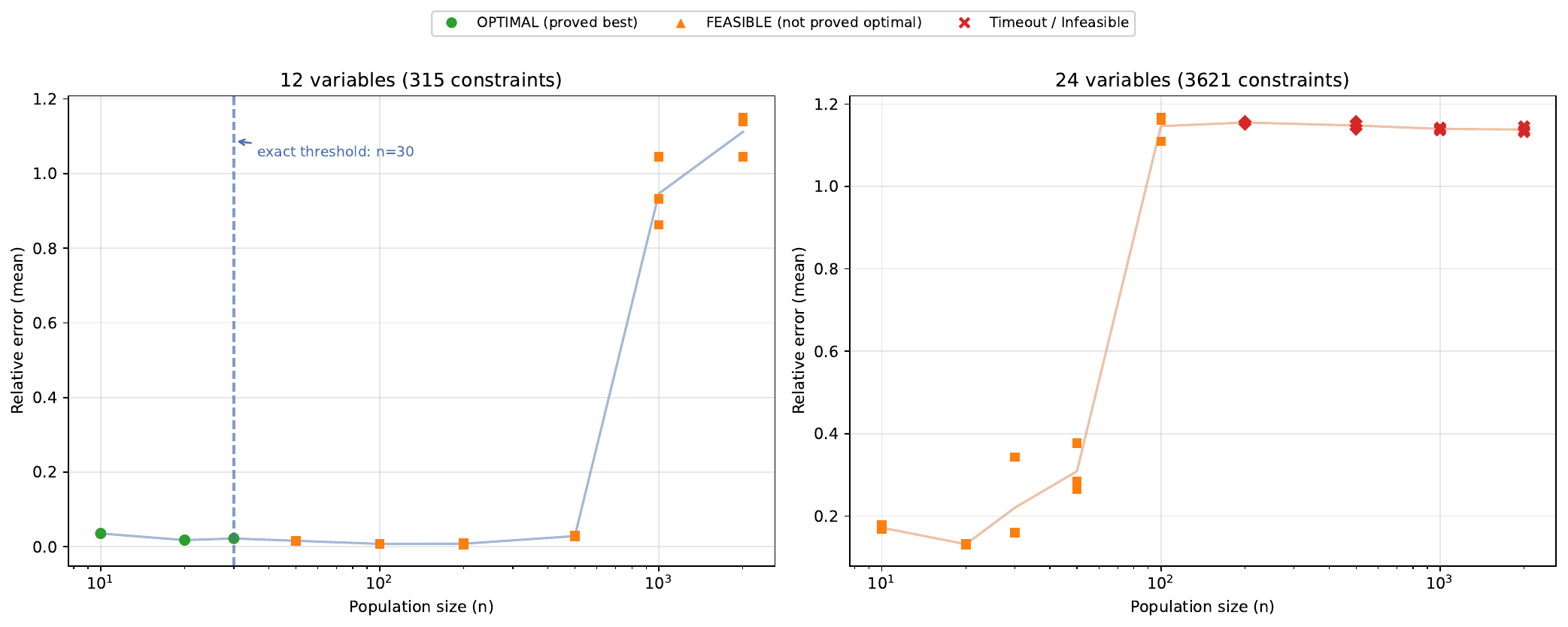}
  \caption{Scalability of the CP-SAT solver. Left: 12 variables (315 target
    cells); the solver proves optimality up to $n = 30$ (green dots). Right:
    24 variables (3{,}621 target cells); the solver never proves optimality.
    Markers indicate solver status: optimal (proved best), feasible (not proved
    optimal), and timeout/infeasible. Error bars show the range over 3 random
    seeds. The time limit is 120\,s per run.}
  \label{fig:cpsat-scalability}
\end{figure}

As expected, the CP-SAT approach computes exact solutions but does not scale.
In the case of 12 variables and 315 target cells, the solver proves optimality
for populations up to $n = 30$ (in under 90 seconds). Beyond this threshold,
the solver returns feasible but unproved solutions whose relative error
degrades rapidly with population size.
In the case of 24 variables and 3{,}621 target cells, the solver cannot prove
optimality even for the smallest tested population ($n = 10$), and fails to
find any feasible solution for $n \geq 200$ within the allotted time.
These results are summarized in Figure~\ref{fig:cpsat-scalability}.

\begin{figure}[H]
\centering
\includegraphics[width=\linewidth]{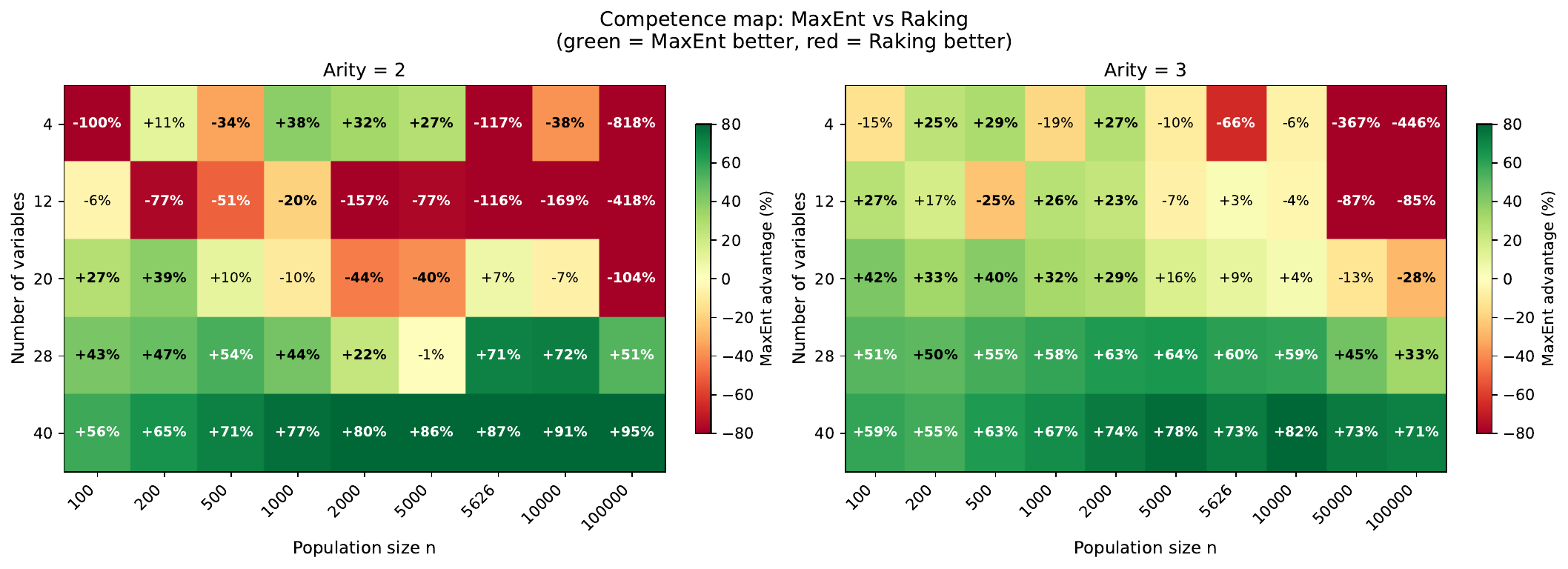}
\caption{Competence map of MaxEnt vs.\ Raking (2D view): relative advantage as a function of population size and number of variables, for arity~2 and arity~3. Green = MaxEnt better; red = Raking better.}
\label{fig:competence-map-2d}
\end{figure}

\begin{figure}[H]
\centering
\includegraphics[width=\linewidth]{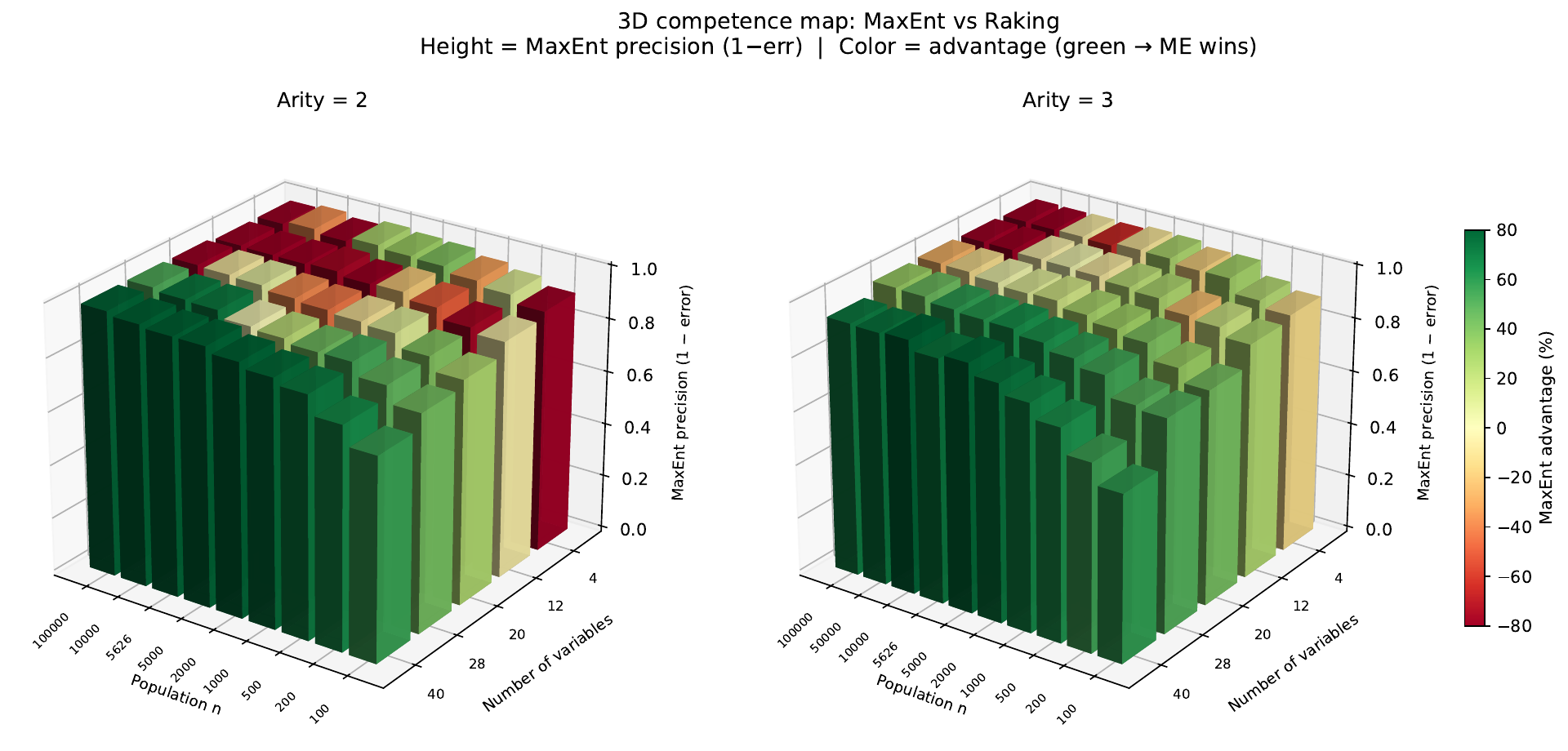}
\caption{Competence map of MaxEnt vs.\ Raking (3D view): height encodes MaxEnt precision (1$-$err); color encodes relative advantage. Green regions indicate configurations where MaxEnt wins.}
\label{fig:competence-map-3d}
\end{figure}

\begin{figure}[H]
\centering
\includegraphics[width=0.48\linewidth]{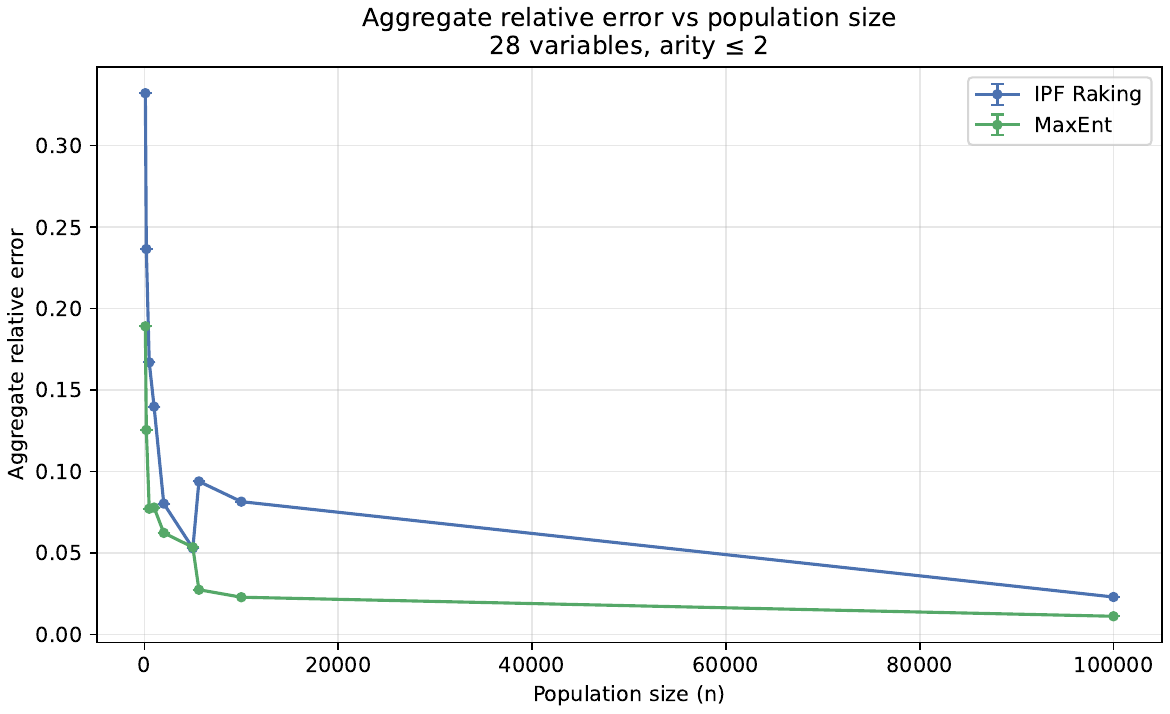}
\hfill
\includegraphics[width=0.48\linewidth]{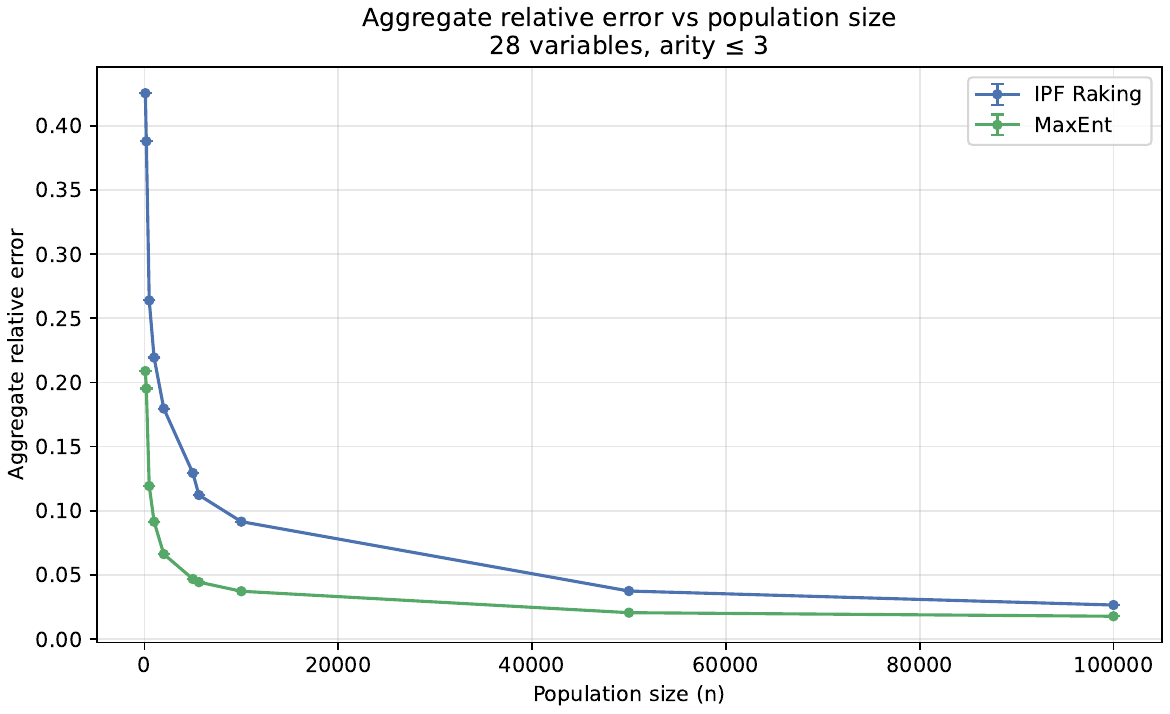}
\caption{Mean relative error vs.\ population size for 28-variable instances (left: arity~2; right: arity~3). MaxEnt consistently outperforms Raking for small populations and high arity.}
\label{fig:error-28var}
\end{figure}

\begin{figure}[H]
\centering
\includegraphics[width=0.48\linewidth]{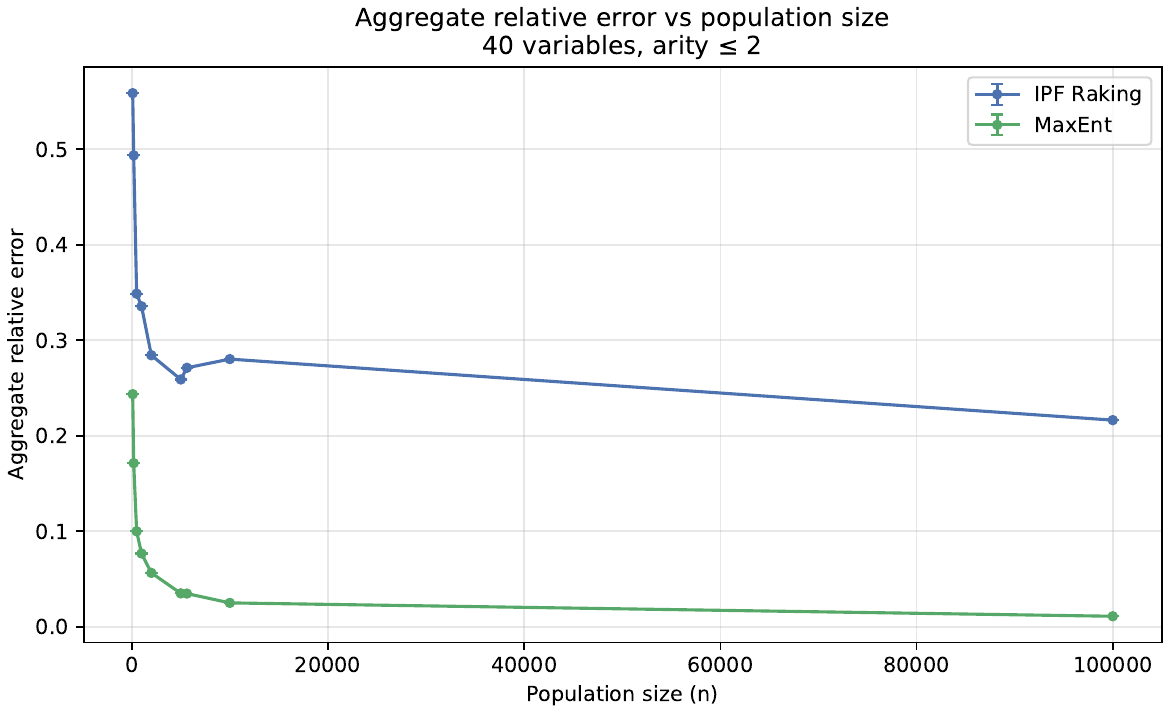}
\hfill
\includegraphics[width=0.48\linewidth]{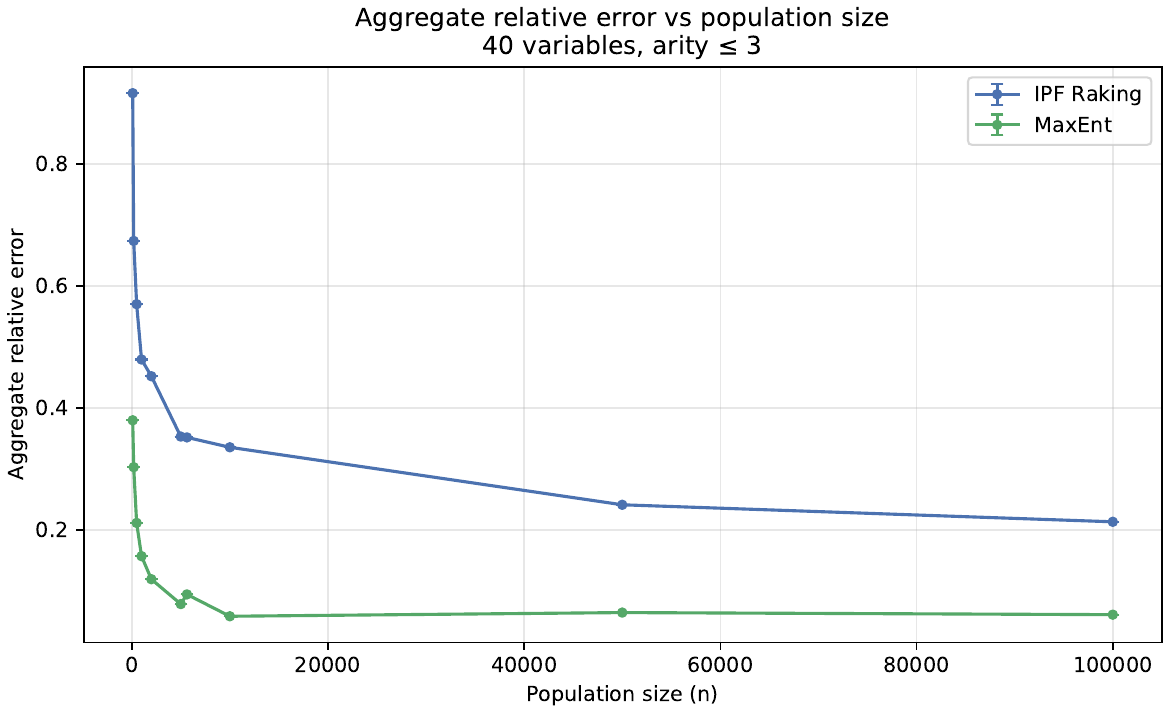}
\caption{Mean relative error vs.\ population size for 40-variable instances (left: arity~2; right: arity~3). The advantage of MaxEnt over Raking grows with the number of variables.}
\label{fig:error-40var}
\end{figure}

Figures~\ref{fig:competence-map-2d} and~\ref{fig:competence-map-3d} summarize the relative performance of MaxEnt and Raking across problem sizes and constraint arities in 2D and 3D competence maps.
Figures~\ref{fig:error-28var} and~\ref{fig:error-40var} show mean relative error as a function of population size for 28- and 40-variable instances, at arity 2 and arity 3.

\section{Discussion}

\subsection{MaxEnt as a Global Optimizer}

From an algorithmic standpoint, the proposed approach can be interpreted as an approximate solver for Sum-based multi-way cardinality CSPs.
Exact CP or MIP formulations search directly for integer-feasible contingency tables.
In contrast, the maximum-entropy relaxation optimizes a single global objective over the multiplier vector $\lambda$ and returns a distribution over complete assignments.
The learned exponential-family model therefore plays the role of a relaxed solver: it identifies regions of the solution space that best satisfy the constraints in aggregate and supports principled sampling rather than arbitrary solution selection.

Crucially, MaxEnt performs a \emph{global} update of all Lagrange multipliers $\lambda$, finding one trade-off across all constraints at once.
Raking, by contrast, applies sequential one-constraint-at-a-time rescalings, so improvements on one marginal can partially undo earlier corrections on overlapping marginals.
On the NPORS-derived benchmarks, this interference becomes more pronounced as $K$ and the constraint arity increase, which helps explain the larger residual errors observed for raking in the high-dimensional settings.

\subsection{Empirical Comparison: MaxEnt vs.\ Raking}

On the NPORS-derived scaling benchmarks, the most salient factor governing relative performance is \emph{the number of variables $K$}, not the sampled population size $N$ alone.
For $K \leq 12$, raking remains a strong baseline and can win at arity~2 or at large $N$ even under arity~3.
Around $K = 20$, the picture becomes mixed: MaxEnt is more reliable at arity~3, whereas neither method dominates uniformly at lower arity.
For $K \geq 28$, MaxEnt outperforms raking in all tested configurations, with the largest gaps appearing on the 40-variable problems.

A plausible structural explanation is that the number of overlapping constraints grows much faster than $N$ as $K$ increases, increasing interference between sequential raking updates while leaving the MaxEnt fit governed by a single global objective.
We present this as an empirical interpretation of the observed results rather than as a general impossibility claim.

Table~\ref{tab:me-vs-raking} reports mean relative constraint error for representative configurations spanning five problem sizes and both arities.

\begin{table}[H]
\centering
\small
\setlength{\tabcolsep}{4pt}
\begin{tabular}{llrrrr}
\toprule
\textbf{Config} & \textbf{Pop.\ $N$} & \textbf{ME} & \textbf{Raking} & \textbf{Winner} & \textbf{Gap} \\
\midrule
\multicolumn{6}{l}{\textit{12 variables}} \\
12-var, arity=2 & 1\,000  & 0.036 & 0.030 & Raking & $-17\%$ \\
12-var, arity=3 & 100    & 0.129 & 0.176 & ME     & $+27\%$ \\
12-var, arity=3 & 5\,626  & 0.026 & 0.026 & equal  & $\approx 0$ \\
12-var, arity=3 & 100\,000& 0.017 & 0.009 & Raking & $-46\%$ \\
\midrule
\multicolumn{6}{l}{\textit{20 variables}} \\
20-var, arity=2 & 200    & 0.114 & 0.186 & ME     & $+39\%$ \\
20-var, arity=3 & 100    & 0.204 & 0.352 & ME     & $+42\%$ \\
20-var, arity=3 & 10\,000 & 0.027 & 0.028 & ME     & $+4\%$ \\
20-var, arity=3 & 100\,000& 0.013 & 0.011 & Raking & $-22\%$ \\
\midrule
\multicolumn{6}{l}{\textit{28 variables — MaxEnt wins in all configurations}} \\
28-var, arity=2 & 1\,000  & 0.078 & 0.140 & ME     & $+44\%$ \\
28-var, arity=3 & 100    & 0.209 & 0.425 & ME     & $+51\%$ \\
28-var, arity=3 & 10\,000 & 0.037 & 0.092 & ME     & $+59\%$ \\
28-var, arity=3 & 100\,000& 0.018 & 0.027 & ME     & $+33\%$ \\
\midrule
\multicolumn{6}{l}{\textit{40 variables — largest observed gaps}} \\
40-var, arity=2 & 1\,000  & 0.077 & 0.336 & ME     & $+77\%$ \\
40-var, arity=3 & 100    & 0.380 & 0.916 & ME     & $+59\%$ \\
40-var, arity=3 & 5\,626  & 0.095 & 0.352 & ME     & $+73\%$ \\
40-var, arity=3 & 100\,000& 0.061 & 0.214 & ME     & $+71\%$ \\
\bottomrule
\end{tabular}
\caption{Mean relative constraint error (lower is better) on representative NPORS-derived scaling benchmarks. Gap = relative reduction over the weaker method.
At $K \leq 12$, the winner depends on arity and $N$. At $K \geq 28$, MaxEnt wins in all configurations we tested.}
\label{tab:me-vs-raking}
\end{table}

\subsection{Empirical Guidance}

The empirical results above support a benchmark-specific rule of thumb, summarised in Table~\ref{tab:recommendations}.

\begin{table}[H]
\centering
\small
\begin{tabular}{lp{5.8cm}}
\toprule
\textbf{Situation} & \textbf{Guidance} \\
\midrule
$K \leq 12$, arity $= 2$                      & Raking is a strong baseline and can be preferred when speed is the main concern. \\
$K \leq 12$, arity $= 3$, small or moderate $N$ & MaxEnt often yields lower error and is worth trying first. \\
$K \approx 20$                                 & Method choice becomes regime-dependent; MaxEnt is safer when ternary interactions matter. \\
$K \geq 28$                                    & MaxEnt consistently outperformed raking on our benchmarks. \\
$K = 40$                                       & The largest empirical gaps favor MaxEnt by a wide margin. \\
\bottomrule
\end{tabular}
\caption{Empirical guidance on the NPORS-derived scaling benchmarks. These recommendations summarize observed trends and should not be read as universal guarantees.}
\label{tab:recommendations}
\end{table}

\section{Conclusion}
We presented a maximum-entropy relaxation of multi-way cardinality-constrained population synthesis.
By interpreting conjunction-defined cardinality requirements as expectation constraints, we obtain an exponential-family model whose parameters can be learned by convex optimization.
On the NPORS-derived scaling benchmarks, MaxEnt is especially attractive when the number of attributes and the density of higher-order interactions increase, whereas generalized raking remains competitive on smaller, lower-arity problems.

Two directions follow naturally from this study.
First, the penalized soft-constraint formulation in \eqref{eq:softmaxent} should be evaluated systematically on genuinely noisy or mutually inconsistent targets.
Second, exact or near-exact repair procedures could be layered on top of MaxEnt samples when selected marginals must be matched exactly.
Extending the same evaluation to higher arities and approximate expectation estimation is an important next step.
\section*{Acknowledgements}
This work originates from research and development activities initiated by the first author in connection with ImagineAllThePeople. The company provided intellectual and conceptual support that significantly contributed to the direction of this work. We thank Mirko Degli Esposti for drawing our attention to an inconsistency between the description of the method in the initial version of this paper and the corresponding implementation.

\bibliographystyle{abbrvnat}
\bibliography{maxent_gcc_ijcai_draft}

\end{document}